# Context-Aware System Synthesis, Task Assignment, and Routing

Jason Ziglar, Ryan Williams, Alfred Wicks

***Abstract*—The design and organization of complex robotic systems traditionally requires laborious trial-and-error processes to ensure both hardware and software components are correctly connected with the resources necessary for computation. This paper presents a novel generalization of the quadratic assignment and routing problem, introducing formalisms for selecting components and interconnections to synthesize a complete system capable of providing some user-defined functionality. By introducing mission context, functional requirements, and modularity directly into the assignment problem, we derive a solution where components are automatically selected and placed into an optimal organization of hardware and software, all while respecting restrictions on component viability and required functionality. The ability to generate *complete* functional systems directly from individual components reduces manual design effort by allowing for a guided exploration of the design space. Additionally, our formulation increases resiliency by quantifying resource margins and enabling adaptation of system structure in response to changing environments, hardware or software failure. The proposed formulation is cast as an integer linear program which is provably $\mathcal{NP}$-hard. Two case studies are analyzed to highlight the expressiveness and complexity of problems that are addressable by this approach: the first explores the iterative development of a ground-based search-and-rescue robot in varying mission contexts, while the second compares an automated humanoid design against a manual design that was finalist in the DARPA Robotics Challenge. Numerical simulations quantify real-world performance and demonstrate tractable time complexity for the scale of problems encountered in many modern robotic systems.**

## I. Introduction

With the popularity of modular robotic software infrastructures such as the Robot Operating System (ROS), the Robot Construction Kit (ROCK), Yet Another Robot Platform (YARP), etc., building novel robotic systems can involve not only developing new hardware and software to generate desired functionality, but also effectively re-using third party development. The availability of multiple components capable of solving a particular technical challenge requires a developer to understand not only what a component nominally provides, but also the *context* in which a component can correctly operate, and which resources are required for operation. As the demand for increasingly complex robotic systems grows, the knowledge of prospective developers must grow exponentially due to the potential interconnections between components and the large range of concerns to track through development. The ecosystem of available software packages eliminates the need for researchers to implement all functionality *de novo*; instead, knowledge of common taxonomies for a given problem, available solutions, and software/hardware requirements helps in solving complex design problems. While reusability reduces development effort, it also introduces a logistical and domain knowledge problem, which can particularly challenge newer researchers discovering an unfamiliar body of knowledge. For instance, in the DARPA Robotics Challenge, robotics groups from across the world competed to build humanoid robots to compete in a series of challenges simulating real world disaster response and rescue scenarios. Team has the option to use pre-built humanoid robots, or fabricate novel robotic platforms. Additionally, teams could use open-source software packages to address topics such as hardware interfacing, walking, localization, obstacle detection, footstep planning, manipulation planning, behavior planning, and user interfaces [1]–[5]. No team deployed a robot using only freely available packages, with many teams instead mixing pre-existing packages with novel research efforts to address competition challenges. A hypothetical team of newcomers attempting to participate would have to survey all necessary problem areas in order to consider available packages, determine their functionality, understand the underlying assumptions, estimate the resources required, and map out how they may integrate with the rest of a possible design, all before deciding whether to use off-the-shelf packages or attempt novel development. Such an analysis would provide information about potential components, but it would not solve the actual challenges in selecting or integrating components into a complete system. The complexity of the space of system designs means these surveys generally focus on qualitative analysis (e.g. whether a component appears to work in a particular context) or empirical evaluation in a largely complete design (e.g. testing if a component works "well enough" within a system). In developing novel systems, this may lead to expending available resources primarily on local improvements or research goals. While such a strategy may render the design space tractable, it fails to rigorously define the design space and provide a complete definition of optimality.

Selecting components to provide some high-level functionality covers only *one* aspect of producing a functional system. Indeed, in producing a complete system a designer must often consider the following issues: (1) components must function correctly in the environment in which a robot operates; (2) hardware must be selected to provide sensing, actuation, and computational resources; (3) software must be assigned the necessary resources; and (4) data must be routed through communication networks without exceeding bandwidth limits. Currently, these tasks are performed largely through manual effort, making complex system design a slow, brittle process. For example, works describing the robots developed in the DARPA Robotics Challenge often include descriptions of the reasoning and testing involved in these manual designs [6], [7], produced over years of effort by large teams. The complex and time-consuming nature of this process typically results in relatively little exploration of the design space, yielding decisions based on expert knowledge and trial-and-error, as opposed to mathematically grounded *optimization*.

This work introduces a method for automatically constructing optimal robotic hardware and software systems from a set of available components, based on a generalization of the quadratic assignment and routing problem. Our formulation provides a unified framework for enabling a wide array of capabilities in the design, development, and operation of robot systems. For design-time operation, our formulation automates the process of selecting components to build a complete system with some user-defined functionality, as well as generating the *structure* that relates all elements (e.g., connecting hardware, assigning tasks, and routing communications). By automating this stage, designs can be made more quickly and with more confidence in the validity of a given solution, since the entire problem is solved simultaneously. This also enables more robust consideration of system resiliency in design, since the impact of small changes in component parameterization (e.g., how much computing resources a particular task needs, the size of a particular message, the cost of using a particular sensor, etc.) can immediately be propagated to the global system design. Furthermore, by fully automating the entire process, system resiliency can be extended by solving the design problem in an *online* setting, through the same process of generating optimal solutions in response to local changes. As examples, changes in the environmental context (e.g., transitioning from indoor to outdoor operation) can require different capabilities (e.g., using GPS for localization), changes in software performance (e.g., a task consuming more resources than anticipated) and changes in hardware components (e.g., computer failure) can require a reallocation of software tasks through the system. The ability to automatically synthesize a novel system capturing these requirements will allow for complex robotic systems that are more efficient and resilient by design.

The main contributions leading to the described formulation are as follows:

1) A formal abstraction defining hardware and software *groups* providing functional capability, which addresses variability in functional decomposition present in state-of-the-art robotic research. It also enables reasoning about a consistent scale of functional definitions, regardless of implementation details.
2) A representation of environmental and contextual requirements for components, yielding systemic and functional requirements that remain constant irrespective of operational context. Contextual requirements for tasks ensure that system synthesis respects the underlying assumptions present in engineered subsystems.
3) A novel set of optimization constraints that capture the structure of both hardware and software composing a robotic system. These constraints unify the synthesis of system structure with assignment and routing, resulting in a tool for understanding how changes at any scale impact an overall system. This can serve to operate as a design-time tool for developing novel robots, as a run-time tool for reconfiguring a system in response to a change in environment, or as a failure response to rebuild a system in case of component failures.
4) An algorithm which operates over this abstraction, and can generate a complete problem formulation.

Two case studies demonstrate the generality and applicability of this approach to a wide range of problems, including heavily-engineered robots. The first case study involves the synthesis of a large number of robotic variants for a search-and-rescue robot operating in a variety of mission contexts, demonstrating the capability of our approach to automate significant portions of an iterative design process. The second case study demonstrates the performance of the proposed approach in synthesizing state-of-the-art robots through the design of ESCHER, a humanoid robot that was manually designed for the DARPA Robotics Challenge. These case studies demonstrate several useful features inherent in our approach, such as automatic dependency resolution, adaptation in response to dynamic mission contexts, and encoding complex realities in mission requirements. Additionally, an empirical study of the scalability of problem instances based on one of the case studies is presented to demonstrate the scale of problems which can be handled.

## II. Related Work

System synthesis is a unified problem capable of addressing several related but traditionally disparate sub-problems. At the application level, the management and assignment of software processes within a robotic system is required, which can be considered a systems engineering problem. At the same time, the assignment of software to hardware and the routing of communication can be formalized as an assignment problem. This provides a method for automatically determining good mappings between a set of tasks and workers, with many useful extensions and generalizations for capturing important details about tasks, workers, and assignments. At a larger scale, the process of assigning jobs to workers can be applied to multi-robot problems, which requires the consideration of dynamic environments, complex interactions between tasks and workers, and constraints due to the physical embedding of task assignments in a team of robots. There exists some work in defining and automating aspects of robot design as well, which are useful in demonstrating the complexity of defining design problems to be amenable to automated approaches.

### A. *Software Infrastructures and Reconfiguration*

Many robotic middleware frameworks include tools to address the run-time aspects inherent in deploying robotic systems. ROS [5] provides tools for specifying system configuration, including the assignment of tasks to multiple computers. This approach involves the operational aspects of managing a complex multi-process software system, easing lifecycle management for systems distributed across a computer network, but does not provide a complete system definition. Model-based representations [8] can specify the overall structure required in a system, as well as the ability for components to meet these requirements, enabling validation of components in meeting overall system requirements. This eases the incremental composition of a system; however these definitions remain in the realm of human experts, as are ensuring non-functional requirements are met. YARP [9] extends these models to ensure software access to specific components such as sensors or specialized computing elements. Tasks can be dynamically assigned based on these resource needs, falling back to a round robin assignment process for tasks without particular needs. This increases flexibility in development and maintenance of a robot as a system evolves, but no guarantee is placed upon the ability to execute, let alone execute in an optimal fashion; instead, the assumption that computational hardware significantly exceeds requirements serves to allow this approach to function. These approaches provide certain aspects of system modeling, definition, and assignment, but do not fully address resource requirements in system design, nor do they support generating structure of hardware and software systems in parallel.

A few proposed software infrastructures have focused on supporting online reconfiguration, which must reason about complete systems. Port-based automatons [10]–[12] provide a framework for reconfiguring software systems for FPGA-based systems in response to online performance metrics. This approach can add, remove, or replace software components while respecting computational limitations, since software can be directly mapped to hardware, but sacrifices more complex constraints such as bandwidth limits or parallel software pipelines (e.g. multiple components using the same data to perform different operations). Other work provides frameworks for expressing higher level models of software components, allowing solutions to be reconfigured or replaced while maintaining synchronization between tasks [13]. These works do not provide for components that represent different decompositions of a set of functional capabilities to be used interchangeably, as we achieve in this work.

### B. Assignment Problems

The theory of combinatorics for task assignment problems has been extensively researched due to applicability in a wide array of domains [14]. Many extensions and variations of the assignment problem exist to capture details of particular applications, starting with the quadratic assignment problem, which introduces flow between assigned tasks [15]. The generalized assignment problem covers assigning multiple tasks to individual agents with budget constraints, allowing varying costs for a given task between different agents [16]. The vector packing problem (or multi-resource extension) represents resources as vectors containing distinct types, capable of representing resource requirements for tasks on complex computers [17]. Routing data through a computer network introduces a second family of problems known as multi-commodity flow problems [18], embedding additional complexity into the overall problem. Most robotic software infrastructures encode transferring data between tasks as an unsplittable flow problem, an assumption which is preserved in our approach. These qualities can be combined into a single problem, resulting in a multi-resource quadratic assignment and routing problem (MRQARP), which captures the case where software and hardware graphs must be specified as *inputs* [19]. MRQARP aims to find the optimal mapping from a given software graph to a hardware graph of computational elements, with the structure of these graphs defined as inputs. Since graph structure is not included in the problem formulation, functional components are not explicitly defined, and the optimization cannot reason over alternatives for a particular element.

### C. Multi-Robot Task Assignment

Task allocation also represents a fundamental building block for collaborative multi-robot systems. In order to achieve high-level autonomous goals and cope with dynamic environments, task allocation models and optimization methods are required that are efficient, scalable, and expressive. Otherwise, allocation plans for multi-robot teams may be intractable or lack sufficient mission complexity. Over the years, a great amount of research has been carried out in the task allocation area within the robotics community. Relevant examples include the sequential auction methods [20]–[23], each solving a variation of the linear assignment problem with provable suboptimality, market-based works [24]–[26] which achieve near-optimal guarantees, combinatoric-based optimization [27], and [28], which provides an early example of abstract task independence through boolean-type relations. It is also important to point out taxonomies that have been performed in task allocation, such as [29] and more recently [30], which provide a far deeper literature survey. The formulation we present in this work can be exploited for multi-robot system synthesis with a greater level of abstraction than is seen in existing multi-robot assignment methods.

### D. Automated Robot Design

Several approaches exist to rigorously define robotic design problems such that they can be automatically solved to produce functional, complete, and viable systems. Early works defined particular system structures to provide models well-suited to applying optimization techniques in deriving parameters to create useful systems [31]. This work demonstrates the utility of powerful system models in automating design, but these models are heavily structured, and do not address adapting system structure to functionality. Defining system design as a *co-design* problem, focusing on selecting components to fulfill subsystem roles to produce optimal designs in the presence of relationships between subsystems provides one such rigorous approach [32], [33]. Co-design allows for reasoning over the interactions in the discrete decisions present in design based on libraries of components, and provides an efficient approach for solving these problems. This approach defines the relationship between subsystems as part of the input for the design problem, requiring manual definition of these relationships which can become cumbersome when combinatoric relationships exist. For instance, solving the problem of routing data through a distributed system would require enumerating all possible routes through a network for all possible messages, as opposed to enumerating all traversable edges for each message, as this approach does. Furthermore, this work directly shows that the problem is fundamentally $\mathcal{NP}$-hard, and by directly tackling the problem as such, leverages the rich literature and available tools for solving such problems. Similarly, tools exist which define the kinematic design of robots in general fashions. Most similar to this work, [34], [35] defines the design as a set of discrete choices to define a robot, representing the selection and interconnection of various components. This approach provides for structure to be understood as a combination of these discrete decisions and a set of rules mapping between these decisions and parameters of underlying design decisions for particular components, providing a framework which can reason about system structure. These approaches focus on connecting system design to the design parameters of components (e.g. the kinematic parameters of an arm, or connecting planning logic to component selection), but do not capture the same level of detail in the low level assignment and routing aspects. Another approach is to start with an initial kinematic design, and define an optimization problem in terms of the same laws of motion in optimizing the design [36]. Starting with an initial design enables defining the design space as an implicit function, enabling the optimization of both discrete and continuous parameters of the kinematic design based on desired functionality. The solution we derive in this work provides greater flexibility in defining elements in terms of potential interactions and limitations, then generates a greater combinatoric expansion of possible designs when selecting an optimal design, including the synthesis of novel structure for organizing components.

## III. The System Synthesis, Task Assignment, and Routing Problem

System synthesis is the problem of building a system capable of executing a set of computational tasks with a user-defined set of functionality. The problem is motivated by a typical development effort encountered when developing robotic systems; experts work from a library of resuable components, and combine them in novel configurations. Components available in these approaches use standardized interfaces which support construction, such as ethernet ports or message passing. An expert selects components, which connections to make between them, and assigns software to hardware to support execution. This approach benefits from enabling reuse of components in new configurations, but is challenged by the fact that many of the underlying choices are well understood but computationally complex problems. This paper formalizes this development approach as the system synthesis problem, and represents the problem as a generalization of the MRQARP.

This problem is broken down into two aspects - the first being a "front end" which provides a method for specifying components which can be as building blocks for systems, and an algorithm for enumerating the possible arrangements these components can take with respect to one another. The second aspect is a "back end" which can find an optimal arrangement of hardware and software, along with an optimal allocation of resources. The front end provides a framework for how components are modeled, and how an overall problem is specified. The back end handles the global optimization of the structural synthesis, task assignment, and data routing. These two aspects are interconnected; functionality often doesn't decompose to the level of granularity the optimization problem addresses, and the back end may not represent a problem in a way that is intuitive to users. This work provides a holistic solution for specifying components and desired designs, and producing optimal solutions. It is important to note that synthesis only produces designs by selecting and arranging components, and does not generate novel elements to introduce into the design.

### A. Specifying Components and Designs

In order to rigorously describe a complex robotic system, we begin by defining a few basic concepts. A *multiset* is a collection of $I$ objects $\Psi = \{\psi_i \mid i = 1, \ldots, I\}$, in which a given object may occur more than once in the collection. The indicator function $\mathbb{1}_{\psi_i}(\Psi) : \Psi \to \mathbb{N}^0$ defines the number of times an object $\psi_i$ occurs in $\Psi$. A *set* is the special case of a multiset in which every object occurs only once, $\mathbb{1}_{\psi_i}(\Psi) = 1 \ \forall \ \psi_i \in \Psi$. A *graph* $\mathcal{G} = (V, E)$ is defined by a set of vertices $V = \{v_i \mid i = 1, \ldots, I\}$, and a multiset of edges $E = \{e_i = \{v_i, v_j\} \mid v_i, v_j \in V\}$ which connect pairs of vertices. The vertices participating in edge $e_i$ are indexed in the form $e_{i,n} \mid n = 1, 2$, also known as a *loop* if $e_{i,1} = e_{i,2}$. Restrictions on the set of edges $E$ define several important classes of graphs which will be used throughout this paper: a *simple graph* possesses a set of edges $E$ in which no edge is a loop; a *multigraph* possesses a multiset $E$ with no loops; and a *pseudograph* possesses a multiset $E$ possibly with loops. Additionally, if the order of vertices in edges is fixed, this defines a *directed* graph, otherwise a graph may be referred to as an *undirected* graph. In this work, graph is used to describe the case in which no assumptions are made about the nature of the graph, with more specific terms used when additional constraints hold true. Furthermore, in order to disambiguate discussions between hardware and software graph elements, hardware vertices and edges are referred to as devices and connections, while software vertices and edges are referred to as tasks and links.

The problem inputs are specified as a set of modules $M$, providing constituent elements and sources of functionality, mission parameters $\vec{r}_s$, and a mission context $\vec{y}_{\tau_p}$. The modules $M = \{\mu_n = \{\Pi_\mu, T_\mu\} \mid n = 1, \ldots, N\}$ defines the library of elements available for synthesis. A module $\mu$ contains a set of tasks $T_\mu = \{\tau_p \mid p = 1, \ldots, P_\mu\}$ and a set of devices $\Pi_\mu = \{\pi_d \mid d = 1, \ldots, D_\mu\}$. A module provides functionality $\vec{c}_\mu = \langle c_{\mu,q} \mid q = 1, \ldots, Q; c_{\mu,q} \in \mathbb{R}^+\rangle$ if included in a design. Missions parameters represent the desired functionality of the system, and are defined by $\vec{r}_s = \langle r_{s,q} \mid q = 1, \ldots, Q; r_q \in \mathbb{R}^+\rangle$. Functionality is assumed to be linearly independent, in the sense that system requirements can be met by summing functionality from included modules. Mission context is defined as $\vec{s} = \langle s_j \mid j = 1,\dots,J; s_j \in \mathbb{R}\rangle$, representing the conditions under which the system operates.

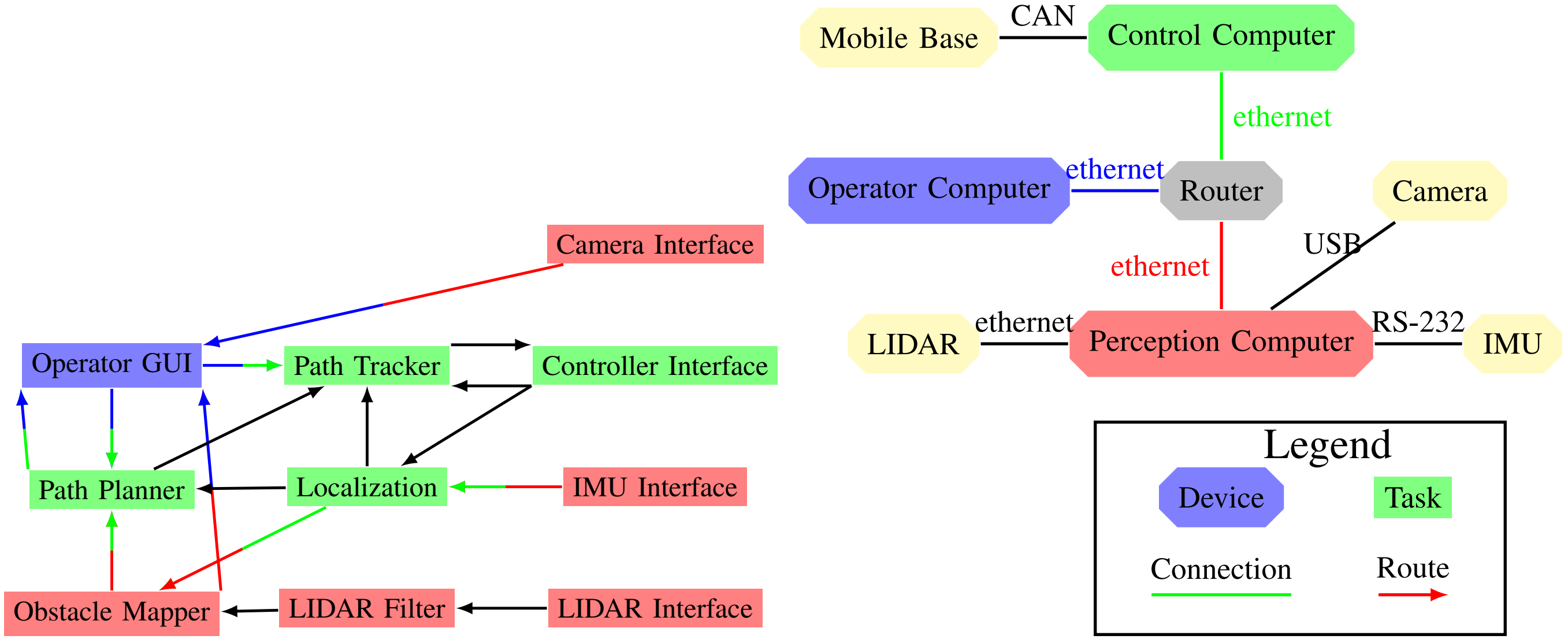

(a) Software graph.

(b) Hardware graph.

Fig. 1. Example of a system synthesis result, containing a hardware graph, software graph, and assignment of software to hardware.Note that in all software graphs, nodes and edges are colorized by the assignment to the hardware graph, and arrows point from the source of data to the sink.

The output of the problem is a computational system $\mathcal{R} = \{\mathcal{H}, \mathcal{S}, \alpha_{\mathcal{H}}^{\mathcal{S}}\}$, comprising of a hardware graph, a software graph, and an assignment of software to hardware. This leads to a full definition of the problem: given a library of modules, construct a computational system providing the mission parameters under the constraints of the mission context. Modules do not explicitly explicitly specify the graphs, but instead provide elements which can be composed into a family of potential designs. These elements are modeled after modular patterns commonly used for constructing systems, such as the publish-subscribe pattern found in ROS, which provide standard paradigms for combining components.

Devices provide physical (e.g. mobility, power, structure) and computational (e.g. CPU cycles, RAM, disk storage) resources to support the operation of the system.

Given $W$ possible resources, a device $\pi_d$ provides $\vec{r}_{\pi_d} = \langle r_{\pi_d,w} \mid w = 1,\dots,W; r_{\pi_d,w} \in \mathbb{R}^+\rangle$ resources. These resources represent computational resources such as available processing power, RAM, disk storage, or other finite resources. Additionally, devices provide some capacity for connecting to other devices over a variety of physical transports. Given $X$ distinct physical transport types, a device provides the capacity to connect to other devices $\vec{\chi}_{\pi_d} = \langle \chi_{\pi_d,x} \mid x = 1,\dots,X; \chi_{\pi_d,x} \in \mathbb{N}^0\rangle$. $\gamma_{x,k}$ represents the $k$-th potential connection using the $x$-th transport type, or alternately $\gamma_x$ if only the transport type is relevant. Connections provide finite bandwidth for transmitting data, resulting in bandwidth limits $B = \{b_{\gamma_x} \mid x = 1,\dots,X; b_{\gamma_x} \in \mathbb{N}^0\}$. Connections also provide resources when connected, represented as $\vec{r}_{\gamma_{x,k}} = \langle r_{x,k,j} \mid j = 1,\dots,J\rangle$, representing logical access to some particular device, such as a particular sensor.

Tasks fulfill computational needs for a system, consuming resources from hardware to accomplish their function. In complex systems, this can include communicating with other tasks to access required data for execution. A task $\tau_p$ consumes resources $\vec{c}^{\tau_p}_{\pi_d} = \langle c^{\tau_p}_{\pi_d,w} \mid w = 1,\dots,W; c^{\tau_p}_{\pi_d,w} \in \mathbb{R}^+\rangle$ when executing on device $\pi_d$. Note that these resources may vary between devices, e.g. due to hardware specialization. A task offers $O_{\tau_p}$ available outputs in the set $\Omega_{\tau_p} = \{\omega_{\tau_p,o} \mid o = 1,\dots,O_{\tau_p}\}$, and requires $I_{\tau_p}$ inputs $\Xi_{\tau_p} = \{\xi_{\tau_p,i} \mid i = 1,\dots,I_{\tau_p}\}$. These outputs and inputs can be linked together to transmit data. Data transmitted in software systems require agreement on typing or structure of data, analogous to physical transports for device connections. We define functions $f_{intype} : \Omega \to \mathbb{N}^0$ and $f_{outtype} : \Xi \to \mathbb{N}^0$, which map outputs and inputs to a unique number for each message type.

Beyond data typing, inputs and outputs in the software multigraph may provide different semantic meaning represent-ing some underlying assumption of a task. For instance, an output transmitting a pose message may contain the estimated pose of a robot, the sensed location of an object of interest, or a commanded goal pose. These instances share a common data type, but represent different quantities in the software system. In order to ensure that the structure keeps semantic consistency, outputs define a vector of $W_{\omega_o}$ resources $\vec{r}_{\omega_o} = \langle r_{\omega_o,w} \mid w = 1, \dots, W_{\omega_o}; r_{\omega_o,w} \in \mathbb{R}^+ \rangle$ which represent the provided semantic output. Similarly, inputs define a vector of $W_{\xi_i}$ requirements $\vec{c}_{\xi_i} = \langle c_{\xi_i,w} \mid w = 1, \dots, W_{\xi_i}; c_{\xi_i,w} \in \mathbb{R}^+ \rangle$ representing the required semantic content for a given input. A task also specifies a required context $\vec{y}_{\tau_p} = \langle y_j \mid j = 1, \dots, J; y_j \in \mathbb{R} \rangle$, representing the underlying assumptions about environment a task encodes.

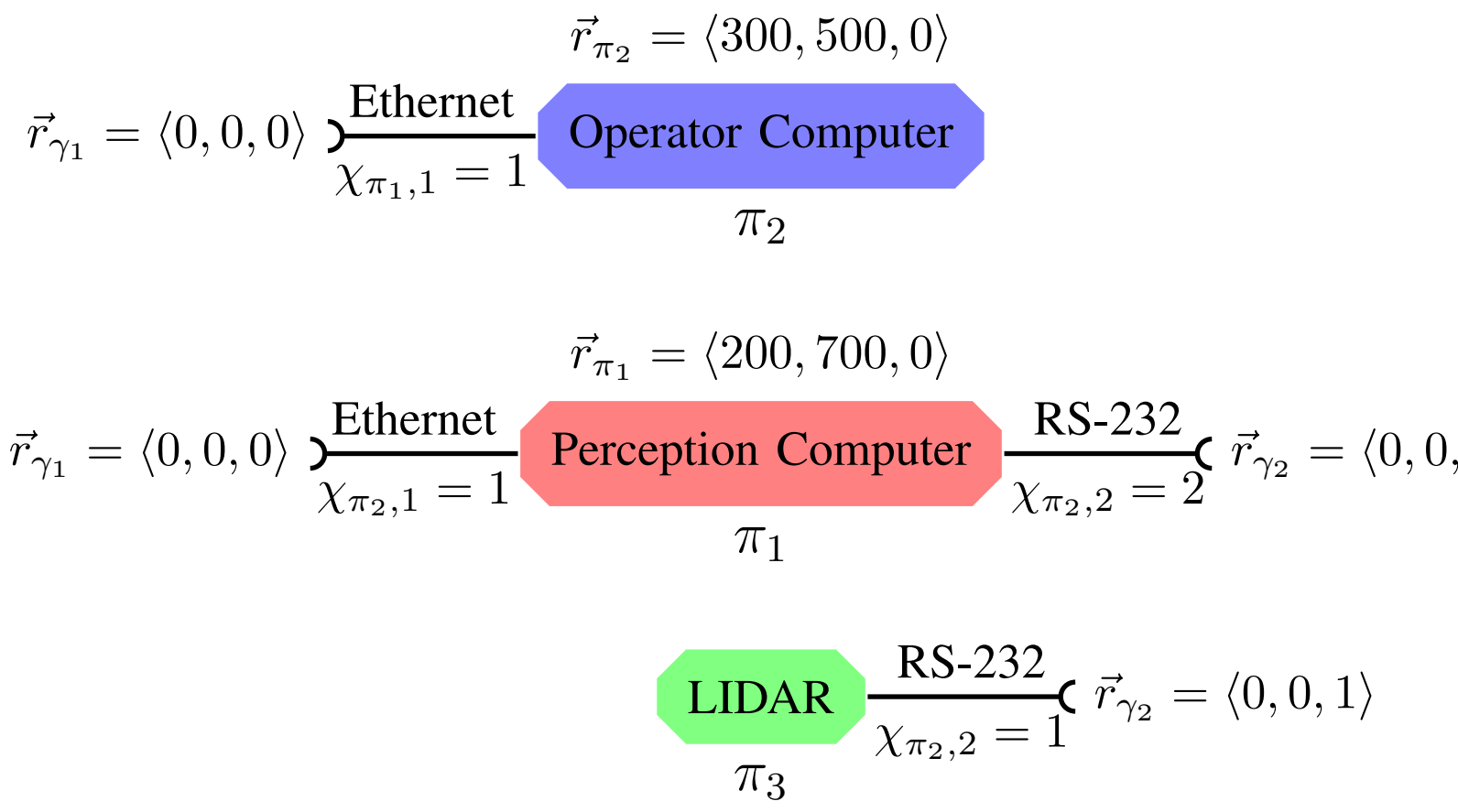

Fig. 2. Example device definitions. Resources are specified as a vector of ⟨CPU, memory, and LIDAR access⟩.

Given a set of devices $\Pi = \{\pi_d \mid d = 1, \dots, D\}$ or tasks $T = \{\tau_p \mid p = 1, \dots, P\}$, it is straightforward to enumerate all possible edges that could make a graph. Both devices and tasks provide a concept of type for their potential connections, and by considering all possible pairs of nodes, the set of potential edges can be enumerated. Given $D$ devices under consideration, the set of possible connections between devices in the pseudograph given a single physical transport can be defined as $\Gamma_x = \{\gamma_{x,k} = \{\pi_u, \pi_v\} \mid k \in E(K_D); \pi_u, \pi_v \in \Pi\}$, where $E(K_D)$ denotes the edges in a complete graph with $D$ vertices. Thus the set of possible connections in the hardware graph is the union of these sets, $\Gamma = \bigcup_{x=1}^{X} \Gamma_x$. Analogously, for each task we can define $\Lambda_{\tau_p} \subset \{\lambda_{\tau_p,\tau_r,\omega_{\tau_p,o},\xi_{\tau_r,i}} \mid \tau_p, \tau_r \in T, \tau_p \neq \tau_r; \omega_{\tau_p,o} \in \Omega_{\tau_p}; \xi_{\tau_r,i} \in \Xi_{\tau_r}\}$ the set of all possible outgoing links from $\tau_p$. All possible links in the software multigraph are then $\Lambda = \bigcup_{\tau_p \in T} \Lambda_{\tau_p} = \{\lambda_l \mid 1, \dots, L\}$. These graphs represent all possible configurations of hardware and software, which allows framing the system synthesis problem as a generalization of the assignment problem, selecting a subset of these two graphs as a form of assignment.

### *B. Structure Synthesis and Resource Allocation*

Previous works accept the hardware and software graphs, $\mathcal{H}$ and $\mathcal{S}$, as the *input* to an optimization problem. A key contribution in this paper is to generalize the MRQARP to accept sets of available hardware $\Pi$ and software $T$ as input, enabling *synthesis* of hardware and software structures defining an optimal computational system $\mathcal{R}$. Additional constraints must be introduced to ensure the graph structure produces a system with two key properties: consistency and viability. Consistency requires that the assignment represents a physically realizable system - devices cannot connect to non-existent devices, tasks cannot communicate data with inactive tasks, and so forth. Viability ensures that the synthesized graphs support the requirements of all constituent elements, while still respecting the assignment and routing constraints described in subsection III-A. For instance, any generated hardware pseudograph must provide the necessary resources to support execution of the software multigraph; devices must have sufficient resources to support task execution, and the connections between devices must provide enough bandwidth for transferring data. Viability addresses only local concerns in generating graphs (e.g. ensuring devices can connect to one another, tasks have required resources and data inputs), deferring treatment of systemic functionality to later constraints. Consistency and viability underlie the novel constraints which enable expanding task assignment and routing to address synthesis.

Selection of a device $\alpha_{\pi_d} \in \mathbb{Z}_2$ represents the decision to include the $d$-th device in the hardware graph. Tasks are assigned to a device $\alpha_{\pi_d}^{\tau_p} \in \mathbb{Z}_2$ to perform computation. Assignments must be binary and atomic, meaning a task $\tau_p$ executes on zero or one $\pi_d \in \Pi$ Equation 3. This assignment must also be consistent, in the sense that a task must execute on a device which is selected Equation 2.

A key observation from using software components from the broader research community is that these components often encode assumptions about the world that constrain their operation. As an example, consider a mission parameter for a hypothetical robot that specifies the ability to localize the robot within the environment. Many possible approaches for robot localization exist [37]–[40], varying in input data, computational complexity, and underlying theory of operation. However, many approaches vary in the assumptions made about the environment, robot, or mission under which the approach will operate; GPS-based localization approaches assume direct visibility of the sky, Kinect-based localization requires indoor environments for the sensor to receive usable data, and 2D localization approaches assume a planar environment. The mission context defines a bounding box within which a task's requirements must exist, generating a contextual constraint in Equation 1.

$$\sum_{\pi_d \in \Pi} \alpha_{\pi_d}^{\tau_p} y_{\tau_d,j} \leq s_j \quad \forall\, \tau_d \in T; j = 1, \dots, J \tag{1}$$

If the mission context includes both outdoor and planar world assumptions, GPS and 2D localization techniques can operate, while indoor-only techniques cannot. When combined with previous constraints, functionality exists when tasks can execute under the correct set of environmental assumptions, with the necessary computational resources, and attached to the appropriate hardware.

With the introduction of inactive devices, tasks can only execute on devices participating in the system (Equation 2), and a link can only route over active connections (Equation 10).

$$\alpha_{\pi_d}^{\tau_p} \leq \alpha_{\pi_d} \quad \forall\, \pi_d \in \Pi; \tau_p \in T \tag{2}$$

The assignment variable $\alpha_{\gamma_{x,k}} \in \mathbb{Z}_2$ selects a particular connection between two devices. Consistent connections respect the fact that a connection can only occur between two selected devices, producing Equation 4. Viable connections must not exceed the connection capacities defined for any device, resulting in Equation 5. Device connections can include a device connecting to itself, covering the case of communication channels such as loopback connections or shared memory communication, resulting in hardware being a pseudograph.

$$\sum_{\pi_d \in \Pi} \alpha_{\pi_d}^{\tau_p} \leq 1 \quad \forall\, \tau_p \in T \tag{3}$$

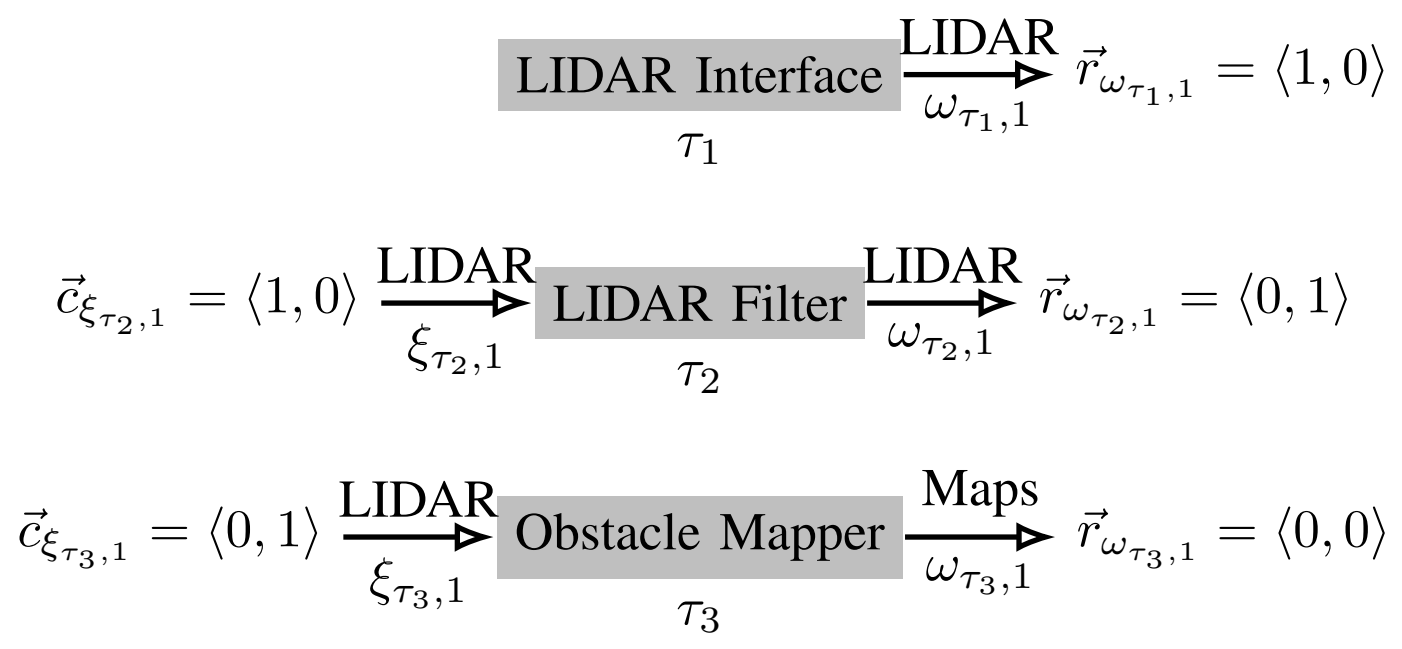

Fig. 3. Example task definitions. All inputs and outputs share the same structure, but their semantic content differs.

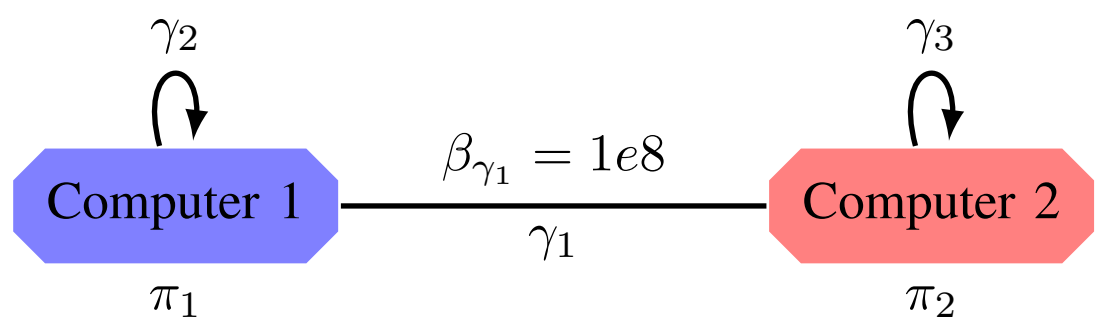

Fig. 4. Simple example of a hardware pseudograph and relevant parameters.

$$\alpha_{\pi_d} \geq \alpha_{\gamma_{x,k}} \quad \forall\, \pi_d \in \Pi; \gamma_{x,k} \in E(\pi_d) \tag{4}$$

$$\sum_{\gamma_{x,k}\in E(\pi_d,x)} \alpha_{\gamma_{x,k}} \leq \chi_{\pi_d,x} \quad \forall\, \pi_d \in \Pi; x = 1,\dots,X \tag{5}$$

A device must provide resources for all assigned tasks, either directly or from active connections, and cannot exceed the available resources Equation 6. For connection provided resources, consider Figure 2, where the LIDAR and the computer executing a device driver are two separate devices. Connecting to the sensor over RS–232 requires selecting the LIDAR and connecting it to the computer over the connection the driver expects. The consumption of computational resources by tasks cannot over-allocate device budgets, resulting in Equation 6.

$$\sum_{\tau_p\in T} \alpha^{\tau_p}_{\pi_d} c^{\tau_p}_{\pi_d,j} \leq r_{\pi_d,j} + \sum_{\gamma_{x,k}\in\Gamma} \alpha_{\gamma_{x,k}} r_{x,k,j} \qquad \forall\, \pi_d \in \Pi; j = 1,\dots,J \tag{6}$$

**Remark.** *The resource provision/consumption constraints required to model hardware requirements are identical in form to those used to represent the provision/consumption of resources for software tasks. This relaxation is not included in this formulation in order to focus on hardware supporting software.*

$$\alpha_{\lambda_l} f_{intype}(\lambda_{l,1}) = \alpha_{\lambda_l} f_{outtype}(\lambda_{l,2}) \quad \forall \lambda_l \in \Lambda \tag{7}$$

Unlike device connections, there exists no fundamental upper limit on the number of links active to a given input, eliminating the need to use a budgetary constraint to limit the number of possible links. While connections between devices provide the capability for transferring data, links between tasks actualize the transfer. For each link, $\alpha_{\lambda_l} \in \mathbb{Z}_2$ represents the decision to transfer data from one task's output to another task's input. All inputs for an active task must have at least one output linked to provide data, but unlike physical connections, no fundamental limit exists on the number of outputs linked to a single input. This allows aggregation of data for processing, leading to a satisfaction constraint in Equation 8 applying to each task input. Consistent routing requires links to only be assigned to active connections Equation 9.

$$\sum_{\pi_d\in\Pi} \alpha^{\tau_p}_{\pi_d} \leq \sum_{\lambda_l\in\Xi_{\tau_p}} \alpha_{\lambda_l} \quad \forall\, \tau_p \in T \tag{8}$$

$$\alpha^{\lambda_l}_{\gamma_k} \geq \alpha_{\lambda_l} \quad \forall\, \lambda_l \in \Lambda; \gamma_k \in \Gamma \tag{9}$$

$$\alpha^{\lambda_l}_{\gamma_k} \leq \alpha_{\gamma_k} \quad \forall\, \gamma_k \in \Gamma; \lambda_l \in \Lambda \tag{10}$$

In addition to active tasks requiring all inputs have at least one assigned link, the assigned links must satisfy the defined budget, as seen in Equation 11. An example of the semantic budget/consumption can be seen in Figure 3, in which all outputs and inputs possess an identical data structure, but semantically different meanings (e.g. raw or ego-filtered data).

$$\sum_{\lambda_l\in\Xi_{\tau_p}} \alpha_{\lambda_l} r_{\lambda_{l,1},j} \geq c_{\xi_i,j} \quad \forall\, \tau_p \in T; \xi_i \in \Xi_{\tau_p}; j = 1,\dots,W_{\xi_i} \tag{11}$$

A link $\lambda_l$ may be assigned to transmit over a connection $\gamma_k$, denoted by the variable $\alpha^{\lambda_l}_{\gamma_k} \in \mathbb{Z}_2$, which consumes the specified amount of bandwidth $c^{\lambda_l}_{\gamma_k}$. Bandwidth consumed can vary due to the differences in connections (e.g., packetized network overhead, requirements on data representations, etc.), requiring the bandwidth utilization to take the connection $\gamma_k$ into account as well. Equation 12 ensures that the assignment of links to connections respects bandwidth limits.

$$\sum_{\lambda_l\in\Lambda} \alpha^{\lambda_l}_{\gamma_k} d^{\lambda_l}_{\gamma_k} \leq b_{\gamma_k} \quad \forall\, \gamma_k \in \Gamma \tag{12}$$

Tasks may be assigned to devices not directly connected to one another, requiring data routing along multiple connections. Multi-hop paths require routing data along a connected path between the devices assigned to each task. Routing is solved using a flow constraint stating that for any device interacting with a link, it must have either an odd number of connections and assigned to the relevant device (i.e., a source or sink) or an even number of connections transmitting the data (i.e., uninvolved or flowing through). This logic is formulated on a per-link basis to produce a linear constraint that scales with the number of edges, instead of the number of paths. The routing must happen over active connections to maintain consistency, which when integrated into a flow constraint results in Equation 13. This introduces a quadratic constraint, which can be linearized by introducing a dummy variable, $\alpha^{\tau_p}_{\varnothing\pi_d} \in \mathbb{Z}_2$, for each task assignment, satisfying when tasks are utilized but not linked, as seen in Equation 14. The linear form of this constraint takes longer to solve in empirical tests, but is included to demonstrate a formulation with purely linear constraints.

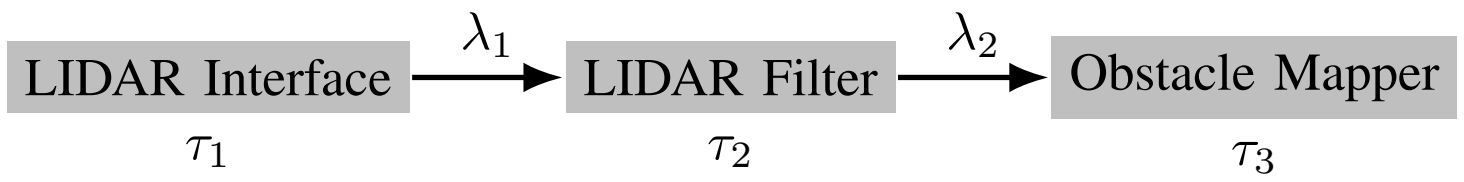

Fig. 5. Example of a software multigraph and relevant parameters.

$$\alpha_{\lambda_l}\Big(\alpha^{\gamma_{k,1}}_{\lambda_{l,1}} + \sum_{\gamma_i \in E(\gamma_{k,1})} \alpha^{\lambda_l}_{\gamma_i}\Big) = \alpha_{\lambda_l}\Big(\alpha^{\gamma_{k,2}}_{\lambda_{l,2}} + \sum_{\gamma_o \in E(\gamma_{k,2})} \alpha^{\lambda_l}_{\gamma_o}\Big) \quad \forall\, \lambda_l \in \Lambda; \gamma_k \in \Gamma \tag{13}$$

$$\frac{\alpha^{\gamma_{k,1}}_{\lambda_{l,1}} + \alpha^{\gamma_{k,1}}_{\varnothing\lambda_{l,1}}}{2} + \sum_{\gamma_i \in E(\gamma_{k,1})} \alpha^{\lambda_l}_{\gamma_i} = \frac{\alpha^{\gamma_{k,2}}_{\lambda_{l,2}} + \alpha^{\gamma_{k,2}}_{\varnothing\lambda_{l,2}}}{2} + \sum_{\gamma_o \in E(\gamma_{k,2})} \alpha^{\lambda_l}_{\gamma_o} \quad \forall\, \lambda_l \in \Lambda; \gamma_k \in \Gamma \tag{14}$$

These constraints ensure selected graph structures maintain consistency in the overall system, and viability during assignment. Combined with the previous constraints for task assignment and routing, any system adhering to these constraints supports the operation of every element.

### C. *Functional Modularity*

The constraints presented thus far provide a method for composing a computational system from a list of available elements, without regard for the overall functionality of the system. Modules provide the framework for introducing functionality to the system synthesis problem, by which inclusion provides functionality. Modules allow an expert to normalize functionality metrics over a set of options, as well as provide any simplifying constraints on the structure of a particular subsystem. These two additions serve to address some of the real world complexities present in determining functional capabilities.

Derived as an instance of an assignment problem, the entire problem has been cast in terms of constraints on individual decisions related to the tasks and devices composing the system. This approach dovetails with the approach taken by many modern robotics software systems, which present tasks as the unit of functionality and reuse [5], [41], [42]. Tasks provide an intuitive unit of functionally reusable software in this context for a variety of reasons: tasks provide a simple model of usability (inputs, process, outputs), require minimal effort to use (launch task, provide data), and can naturally isolate development of novel algorithms. Tying reusability to an individual executable black box presents a problem when considered across an ecosystem consisting of diverse contributions. Reusability tends to favor smaller units of modularity [43], exerting downward pressure to produce tasks with compact sets of functionality. However, the definition of functional completeness for an individual task varies between different approaches, depending on performance considerations, underlying algorithms, or philosophical views. Tasks define functional scope differently depending on the individual resolution of this tension, breaking the direct coupling between task selection and functional capability. Depending on the decomposition of functionality in a given problem, some units of functionality will inevitably require more than a single task to implement. In order to introduce functionality requirements, this work introduces a mid-level concept of *functional modules*.

A module represents a collection of tasks and devices which, if included, provide some amount of functionality to the overall system. A module only functions when all elements are included, introducing an atomicity relationship between all elements and each individual task (Equation 15) and device (Equation 16). While a new assignment variable, $\alpha_\mu$, could be introduced to indicate the selection of a module, this variable can be represented in terms of the assignment of component elements. To minimize the number of variables in the formulation, the indicator function in Equation 17 will be used, although $\alpha_\mu$ will be used for brevity. The sum of the functionality vectors of active modules must meet or exceed the mission parameters $\vec{r}_s$, resulting in the final constraint on mission readiness in Equation 18.

$$(D_\mu + P_\mu)\sum_{\pi_d \in \Pi} \alpha^{\tau_p}_{\pi_d} = \sum_{\pi_d \in \Pi_\mu} \alpha_{\pi_d} + \sum_{\tau_r \in T_\mu}\sum_{\pi_d \in \Pi} \alpha^{\tau_r}_{\pi_d} \quad \forall\, \mu \in M; \tau_p \in T_\mu \tag{15}$$

$$(D_\mu + P_\mu)\alpha_{\pi_d} = \sum_{\pi_r \in \Pi_m} \alpha_{\pi_r} + \sum_{\tau_d \in T_m}\sum_{\pi_r \in \Pi} \alpha^{\tau_d}_{\pi_r} \quad \forall\, \mu \in M; \pi_d \in \Pi_\mu \tag{16}$$

$$\alpha_\mu = \frac{\sum_{\pi_d \in \Pi_\mu} \alpha_{\pi_d} + \sum_{\tau_d \in T_\mu}\sum_{\pi_d \in \Pi} \alpha^{\tau_p}_{\pi_d}}{D_\mu + P_\mu} \tag{17}$$

$$\sum_{\mu \in M} \alpha_\mu c_{\mu,q} \geq r_{s,q} \quad \forall\, q = 1, \ldots, Q \tag{18}$$

Comparing functionality metrics directly at the module level enables reasoning about the opportunity costs in a design,

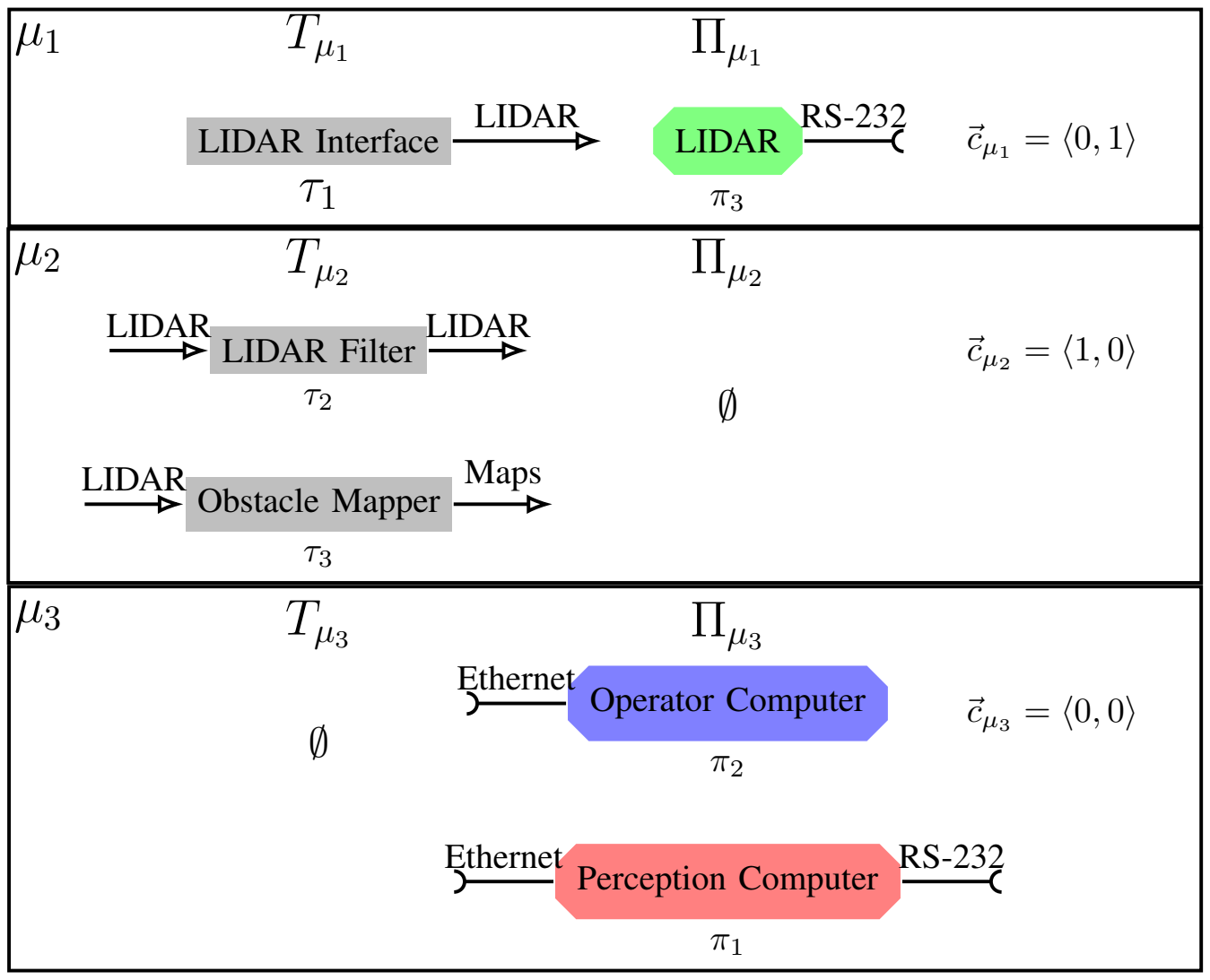

Fig. 6. Example tasks organized into modules. Requiring $\vec{c}_{\mu_2}$ from $\mu_2$ requires all three modules to fully satisfy constraints.

since assignment can reason about the tradeoffs in selecting one module over another, similar to a trade study performed by an expert. Crucially, modules only contain elements directly related to the defined functional capabilities, but not necessarily all elements required to produce a complete system. This functional independence, combined with previous constraints, introduces automatic dependency resolution among modules - selecting one module may require the selection of other modules to produce a complete system. The variable scale of modules and functionality capability in modules, combined with the assurance that the resulting system ensures a complete system, allows for applying this approach to a wide range of system synthesis problems. Modules can represent small, single function components for a single robot as easily as they can represent individual pre-built robots composing a complex multi-robot system, serving to unify design for a wide range of robotic systems.

### D. *Objective Function and Full Formulation*

As a constrained minimization problem, combining the constraints defined previously with an objective function completes the formulation for selecting an optimal system. These cost functions can be defined in multiple ways, depending on the desired metrics to define an optimal system. The cost function is presented in a generalized fashion to enable a point by which expert knowledge can be applied to define optimality for a given scenario. This work decomposes the objective function into five parts: modules, devices, execution, connections, and routing. The objective function in Equation 20 composes the overall cost function from terms representing the user defined costs for each selection.

Inclusion of a module itself may have some cost $f_o : M \rightarrow \mathbb{R}^+$. Similarly, including a device induces a cost $f_{dev} : \Pi \rightarrow \mathbb{R}^+$. For brevity, these costs are combined into a single module cost, since they both can be included by the same assignment decision Equation 19. Somewhat analogously, the cost of including a given task is defined as a function $f_{exec} : \{\Pi, T\} \rightarrow \mathbb{R}^+$, though this may include both the cost of execution (e.g. resource consumption) and including the task at all (e.g. license fees). This cost is not included in the module cost due to being a function of the assignment to a specific device, even though tasks are part of modules.

$$f_m = (f_o(\mu) + \sum_{\pi_d \in \Pi_\mu} f_{dev}(\pi_d)) \tag{19}$$

Device connection costs of the form $f_{cnx} : \Gamma \rightarrow \mathbb{R}^+$ not only serve the previously stated system goals, but also serve to handle physicality constraints on system components. Assigning infinite costs to physical connections between devices (e.g. wired connections) ensures that connections between devices cannot impose unwanted physical connections. Common cases for this need include remote operator stations, where a robot needs to transmit data without constraining motion, or multi-robot systems, in which individual robots must remain physically independent. Message routing introduces some cost $f_{route} : \{\Gamma, \Lambda\} \rightarrow \mathbb{R}^+$ for utilizing bandwidth available on device connections, which aims to reduce latency. The particulars of modeling latency in networks are beyond the scope of this problem formulation, but instead, the intuition that an oversubscribed connection will result in delivery time increasing unbounded serves to motivate that reducing bandwidth utilization will minimize latency. The relationship between bandwidth utilization and latency can vary depending on the devices connected, the physical transport, and the link routed over a particular connection. Note that while these two costs are logically separate, the cost function in Equation 20 combines them in a slightly more concise form, which includes the message routing cost in the device connection equation. These functions are combined to produce the total system cost function in Equation 20a.

The selection of sub-functions not only defines the objective function, but also defines the complexity of the given problem. If all functions $f$ are linear, the final problem is an integer linear program; if any function is quadratic, the problem is instead a quadratic program. As an example, task costs can be quantified by linear metrics (e.g. resource utilization) or quadratic metrics (e.g. load balancing, power utilization, thermal load). In order to demonstrate the overall capability of system synthesis to construct optimal systems, the remainder of this work will assume linear functions, and thus an integer linear program. While many other frameworks exist for solving this type of constrained optimization (e.g. guided local search, genetic algorithms, simulated annealing), an integer linear program can be solved for the global optimum, guaranteeing an optimal solution and upper bound on performance. It is important to note that this is a generalization of MRQARP [19], which has been proven to be $\mathcal{NP}$-hard in the strong sense [44]. The additional variables representing the structure of the graphs and functional requirements can be set to trivial values such that a problem instance maps to an instance of MRQARP, leading this problem to share the same complexity class.

$$Z = \texttt{min} \sum_{\mu \in M} \left( \alpha_\mu \left( f_o(\mu) + \sum_{\pi_d \in \Pi_\mu} f_{dev}(\pi_d) \right) \right) + \sum_{\tau_p \in T} \sum_{\pi_d \in \Pi} \alpha_{\pi_d}^{\tau_d} f_{exec}(\pi_d, \tau_p) + \sum_{\gamma_k \in \Gamma} \left( \alpha_{\gamma_k} f_{cnx}(\gamma_k) + \sum_{\lambda_l \in \Lambda} \alpha_{\gamma_k}^{\lambda_l} f_{route}(\gamma_k, \lambda_l) \right) \tag{20a}$$

s.t.

$$\sum_{\mu \in M} \alpha_\mu c_{\mu,q} \geq r_{s,q} \quad \forall\, q = 1, \ldots, Q \tag{20b}$$

$$\sum_{\pi_d \in \Pi} \alpha_{\pi_d}^{\tau_p} y_{\tau_d,j} \leq s_j \quad \forall\, \tau_d \in T; j = 1, \ldots, J \tag{20c}$$

$$(D_\mu + P_\mu) \sum_{\pi_d \in \Pi} \alpha_{\pi_d}^{\tau_p} = \sum_{\pi_d \in \Pi_\mu} \alpha_{\pi_d} + \sum_{\tau_r \in T_\mu} \sum_{\pi_d \in \Pi} \alpha_{\pi_d}^{\tau_r} \quad \forall\, \mu \in M; \tau_p \in T_\mu \tag{20d}$$

$$(D_\mu + P_\mu) \alpha_{\pi_d} = \sum_{\pi_r \in \Pi_\mu} \alpha_{\pi_r} + \sum_{\tau_d \in T_\mu} \sum_{\pi_r \in \Pi} \alpha_{\pi_r}^{\tau_d} \quad \forall\, \mu \in M; \pi_d \in \Pi_\mu \tag{20e}$$

$$\alpha_{\gamma_k}^{\lambda_l} \leq \alpha_{\gamma_k} \quad \forall\, \gamma_k \in \Gamma; \lambda_l \in \Lambda \tag{20f}$$

$$\alpha_{\pi_d}^{\tau_p} \leq \alpha_{\pi_d} \quad \forall\, \pi_d \in \Pi, \tau_p \in T \tag{20g}$$

$$\alpha_{\lambda_l} f_{intype}(\lambda_{l,1}) = \alpha_{\lambda_l} f_{outtype}(\lambda_{l,2}) \quad \forall \lambda_l \in \Lambda \tag{20h}$$

$$\alpha_{\lambda_l}(\alpha_{\lambda_{l,1}}^{\gamma_{k,1}} + \sum_{\gamma_i \in E(\gamma_{k,1})} \alpha_{\gamma_i}^{\lambda_l}) = \alpha_{\lambda_l}(\alpha_{\lambda_{l,2}}^{\gamma_{k,2}} + \sum_{\gamma_o \in E(\gamma_{k,2})} \alpha_{\gamma_o}^{\lambda_l}) \quad \forall\, \lambda_l \in \Lambda; \gamma_k \in \Gamma \tag{20i}$$

$$\sum_{\lambda_l \in E^-(\tau_p)} \alpha_{\lambda_l} r_{\lambda_{l,1},j} \geq c_{\xi_i,j} \quad \forall\, \tau_p \in T; \xi_i \in \Xi_{\tau_p}; j = 1, \ldots, W_{\xi_i} \tag{20j}$$

$$\sum_{\pi_d \in \Pi} \alpha_{\pi_d}^{\tau_p} \leq \sum_{\lambda_l \in \Xi_{\tau_p}} \alpha_{\lambda_l} \quad \forall\, \tau_p \in T \tag{20k}$$

$$\alpha_{\gamma_k}^{\lambda_l} \geq \alpha_{\lambda_l} \quad \forall\, \lambda_l \in \Lambda; \gamma_k \in \Gamma \tag{20l}$$

$$\sum_{\pi_d \in \Pi} \alpha_{\pi_d}^{\tau_p} \leq 1 \quad \forall\, \tau_p \in T \tag{20m}$$

$$\sum_{\tau_p \in T} \alpha_{\pi_d}^{\tau_p} c_{\pi_d,j}^{\tau_p} \leq r_{\pi_d,j} + \sum_{\gamma_{x,k} \in \Gamma} \alpha_{\gamma_{x,k}} \gamma_{x,k,j} \quad \forall\, \pi_d \in \Pi; j = 1, \ldots, J \tag{20n}$$

$$\alpha_{\pi_d} \geq \alpha_{\gamma_{x,k}} \quad \forall\, \pi_d \in \Pi; \gamma_{x,k} \in E(\pi_d) \tag{20o}$$

$$\sum_{\gamma_{x,k} \in E(\pi_d,x)} \alpha_{\gamma_{x,k}} \leq \chi_{\pi_d,x} \quad \forall\, \pi_d \in \Pi; x = 1, \ldots, X \tag{20p}$$

$$\sum_{\lambda_l \in \Lambda} \alpha_{\gamma_k}^{\lambda_l} d_{\gamma_k}^{\lambda_l} \leq b_{\gamma_k} \quad \forall\, \gamma_k \in \Gamma \tag{20q}$$

$$\alpha_{\pi_d}, \alpha_{\pi_d}^{\tau_p}, \alpha_{\lambda_l}, \alpha_{\gamma_k}^{\lambda_l} \in \mathbb{Z}_2 \quad \forall\, \pi_d \in \Pi; \tau_p \in T; \gamma_k \in \Gamma; \lambda_l \in \Lambda \tag{20r}$$

### E. Problem Generation

Given the full formulation of the system synthesis problem, the complete problem defines hardware and software graphs over all possible combinations of both hardware and software, as well as all possible assignments of software to hardware. Manually enumerating all possible combinations to fully define a problem is prohibitive, and does not match how common ecosystems define these problems. Instead, systems are defined by rules for how components can be combined, and can then provide components which adhere to these rules. Since this composes the problem based on descriptions of elements, tasks and devices can be defined independent of the overall system. This allows for more compact definitions as well as re-use of element definitions when appropriate (e.g. common computers). Following similar logic as used to define the complete graph of the system synthesis problem, Algorithm 1 takes a library of components, and automatically generates a complete instance of the system synthesis problem. This algorithm allows for a compact and intuitive representation of the system synthesis problem, with the full combinatoric complexity available in modern, modular frameworks (such as the Robotic Operating System) evaluated without manual effort.

## IV. Simulation Results

To demonstrate applicability and performance of the system synthesis, two case studies are presented and benchmarked. The case studies represent two traditionally different stages of building robotic systems: the incremental design and development of a robot, and the construction of a complex fielded robot. These problems demonstrate the expressiveness and capability of system synthesis, and provide some insight into performance with realistic systems.

The first case explores the power of context and functional requirements in synthesizing a complex system and simplified variants in the face of differing contexts. This case considers the development of a single ground-based robot for a search and rescue mission. As a hypothetical research system, components are developed in isolation, and integrated at some later point. Integration exercises several potential complexities for system synthesis: components need to share resources, subsystem testing may occur under differing contexts, and modules which indirectly support functional requirements (e.g. adapting interfaces between two tasks, providing additional computing) may be required to enable integration. This experiment considers a fixed set of modules, demonstrating the

**Algorithm 1** Generate System Synthesis Program

**procedure** GENERATE($M, \vec{r}_s, \vec{c}_m, f_o, f_{dev}$)
  $\Pi, T, \Gamma, \Lambda, cnx_limits, constr \leftarrow \{\}$
  $cost \leftarrow 0$
  **for** $\mu \in M$ **do**
    $\Pi \leftarrow \Pi \cup \Pi_\mu$
    $T \leftarrow T \cup T_\mu$
    $cost \leftarrow cost + \alpha_\mu \left( f_o(\mu) + \sum_{\pi_d \in \Pi_\mu} f_{dev}(\pi_d) \right)$
    $constr \leftarrow constr \cup$ AtomicModule($\mu$) $\cup$ ModuleFunctionality($\mu, \vec{r}_s$)
  **end for**
  **for** $\pi_d, \tau_p \in \Pi \times T$ **do**
    $cost \leftarrow cost + \alpha_{\pi_d,\tau_p} f_{exec}(\pi_d, \tau_d)$
    $constr \leftarrow constr \cup$ AtomicTask($\tau_p$) $\cup$ InBudget($\pi_d, \tau_p$) $\cup$ ExecOnActive($\pi_d, \tau_p$)
  **end for**
  **for** $\pi_1, \pi_2 \in \binom{\Pi}{2}; x = 1, \ldots, X$ **do**
    **if** $min(\chi_{\pi_1,x}, \chi_{\pi_2,x}) > 0$ **then**
      $\gamma_k \leftarrow \{\pi_1, \pi_2\}$
      $\Gamma \leftarrow \Gamma \cup \{\gamma_k\}$
      $cnx_limits(\pi_1, x) \leftarrow cnx_limits(\pi_1, x) + 1$
      $cnx_limits(\pi_2, x) \leftarrow cnx_limits(\pi_2, x) + 1$
      $constr \leftarrow constr \cup$ ActiveCnx($\pi_1, \pi_2$)
      $cost \leftarrow cost + \alpha_{\gamma_k} f_{cnx}(\gamma_k)$
    **end if**
  **end for**
  **for** $\pi_d \in \Pi; x = 1, \ldots, X$ **do**
    $constr \leftarrow constr \cup$ CnxCapacity($\pi_d, x, cnx_limits(\pi_d, x)$))
  **end for**
  **for** $\tau_1, \tau_2 \in$ PERMUTE($T$, 2) **do**
    **for** $(out, in) \in$ GetOutputs($\tau_1$) $\times$ GetInputs($\tau_2$) **do**
      **if** $type(out) == type(in)$ **then**
        $\Lambda \leftarrow \Lambda \cup \{\{\tau_1, \tau_2\}\}$
        $constr \leftarrow constr \cup$ ActiveInputs($in$) $\cup$ LinkResources($in, out$)
      **end if**
    **end for**
  **end for**
  **for** $\gamma_k, \lambda_n \in \Gamma \times \Lambda$ **do**
    $constr \leftarrow constr \cup$RouteOnActive($\gamma_k, \lambda_n$)$\cup$ BandwidthLimit($\gamma_k, \lambda_n$)$\cup$Flow($\gamma_k, \lambda_k$)
    $cost \leftarrow cost + \alpha_{\gamma_k,\lambda_n} f_{route}(\gamma_k, \lambda_n)$
  **end for**
  **return** $cost, constr$
**end procedure**

versatility in resulting designs as a result of evolving functional integration.

The second case study considers the design of a humanoid robot for the DARPA Robotics Challenge. The development of the custom humanoid ESCHER [6] complex enough to require a large team of engineers to handle the same challenges presented in this framework, and familiar enough to the authors to formulate as a system synthesis problem. The system requires a diverse set of hardware to support the operation of a custom 33 degree-of-freedom, force controlled robot, as well as a large number of software tasks to provide a sliding autonomy framework, as well as handling degraded communications. As a robot fielded without the use of the proposed framework, significant manpower went into manually defining and validating the overall system design, providing a natural comparison for the optimization. This provides both a comparison against the state-of-the-art in designing these systems, as well as a demonstration of applicability to real world problems.

In the formulation, the concrete definition of the cost function is deferred to be problem or user specific. For the following experiments, the same set of cost functions are used to compose the objective function, and are defined here for clarity. Task execution costs are defined as the fractional consumption of resources (Equation 21), normalizing over scales of computational resources and favors parsimonious systems. Route costs define the cost of traversing a loop as zero cost, but otherwise uses the same fractional consumption cost as task execution (Equation 22). Connection costs serve only to disallow physical connections between systems which cannot support physical connections in the problem context (e.g. a mobile robot and a remote operator), otherwise providing a constant cost for any connection. Module and device costs are also defined in a problem specific manner, estimating the monetary cost for purchasing the device new (or the module, in the case of modules representing complex devices.) These costs aim to define optimal systems as ones which minimize the overall cost of constructing the final system, by minimizing the amount of computational and bandwidth resources required, and then considering monetary costs.

$$f_{exec}(\pi_d, \tau_p) = \sum_{w=1}^{W} \frac{c_{\pi_d,\tau_p,w}}{r_{\pi_d,w}} \tag{21}$$

$$f_{route}(\gamma_k, \lambda_n) = \begin{cases} 0 : \gamma_{k,1} = \gamma_{k,2} \\ \frac{c^{\lambda_l}_{\gamma_{k,n}}}{b_{\gamma_k}} : \gamma_{k,1} \neq \gamma_{k,2} \end{cases} \tag{22}$$

Besides the cost function, all instances are generated by Algorithm 1, with the specific element parameters (e.g. computational resources) empirically derived. Problem formulations generated in this fashion are solved using the Gurobi [45] optimization library.

### A. Dynamic Robot Development

Consider the development of a mobile robotic system for tracking and retrieving a potential target of interest. The robot requires a few elements for operation: a method for finding and localizing a target of interest, a manipulator for grasping a target, and a method for selecting grasp poses. Component development and testing initially occurs in isolation, in environments which do not fully replicate operational conditions. The devices and tasks defined for this system are derived from real-world robots developed for this task, using a ROS-based system for providing many of the components defined. For this problem, the context is defined with three binary dimensions: the time of operation (day or night), location (indoors or outdoors), and whether the targets of interest are visually identifiable or contain an active beacon. These contextual parameters limit the applicability of some defined options: target tracking options include an RGB camera tracking system (capable of working in daytime scenarios), a monochrome night-vision system (capable of working at night for visible targets), and a system for triangulating a signal from a non-visual target (works regardless of lighting, but at a reduced tracking range.) Localization modules include GPS for absolute positioning outdoors and a variety of simultaneous localization and mapping (SLAM) packages making different tradeoffs for computational effort and positional accuracy. Functional requirements cover both required hardware capability (e.g. a mobile base for traversing terrain, an arm for manipulating a target,) required software functionality (e.g. target tracking,) and differing levels of autonomy (e.g. teleoperated or planning for the mobile base or arm, or if planning should happen completely autonomously), and include both binary requirements (e.g. presence of absence of an arm) and continuous parameters (e.g. detection range for targets). The options and variability expressed in this problem provide some of the uncertainty and variability present in real world development projects.

Given this problem, the versatility of system synthesis in adapting to requirements can be demonstrated. We define a fixed set of functional modules ($D = 19, P = 25, N = 29$) covering the available options for the entire range of capability described above. This includes several computing, sensing, localization, and tracking modules, with multiple options provided for many aspects. These modules providing the described options are based on off-the-shelf components available for solving relevant problems, such as GMapping [46], Hector SLAM [47], and commercially available hardware. These modules are then combined with each valid combination of contextual parameters and functional requirements to generate a full range of variant problem instances. Table I reports the mean solution time for three trials of each full problem configuration.

In order to analyze the impact of functional requirements and contextual parameters, an ANOVA test is performed with results reported in [48]. The results generally indicate that parameters which impact trade-offs significantly alter the computation time, as opposed to parameters which include or exclude static subsystems. Including a manipulator as a functional requirement introduces complexity in the final system in terms of the computational system, but is orthogonal to contextual considerations. Requiring target tracking introduces consideration of three possible options, which operate under different contexts, introduces significant changes to computation time. Functional requirements which implicitly include contextual interactions, such as the mobile base (which requires localization to operate, linked to context) also respond accordingly. Of the interactions considered, the impact of both the arm and target tracker as individual factors are masked by the interaction between these components. Contextual parameters, which interact with all trade-off considerations in the system, impact computation time as well. While significant, these parameters introduce relatively small changes in the overall time, requiring at most $0.341s$ additional time.

TABLE I

Mean solution times (seconds) for different configuration and contexts

For contexts, D=Day, N=Night, O=Outdoors, I=Indoors, V=Visual, B=Beacon

| Capability | DIB | DIV | DOB | DOV | NIB | NIV | NOB | NOV |
|---|---|---|---|---|---|---|---|---|
| Track | 0.322 | 0.313 | 0.216 | 0.257 | 0.251 | 0.289 | 0.216 | 0.241 |
| Drive | 0.244 | 0.298 | 0.250 | 0.256 | 0.253 | 0.265 | 0.246 | 0.234 |
| Arm | 0.228 | 0.268 | 0.239 | 0.231 | 0.304 | 0.244 | 0.240 | 0.218 |
| Track, Drive | 0.245 | 0.299 | 0.246 | 0.249 | 0.250 | 0.271 | 0.250 | 0.234 |
| Track, Arm | 0.249 | 0.312 | 0.214 | 0.262 | 0.253 | 0.289 | 0.215 | 0.241 |
| Drive, Plan | 0.462 | 0.560 | 0.219 | 0.273 | 0.478 | 0.5169 | 0.219 | 0.244 |
| Arm, Plan | 0.298 | 0.317 | 0.215 | 0.264 | 0.303 | 0.343 | 0.217 | 0.235 |
| Track, Drive, Plan | 0.472 | 0.562 | 0.220 | 0.271 | 0.471 | 0.553 | 0.218 | 0.249 |
| Track, Arm, Plan | 0.260 | 0.319 | 0.228 | 0.268 | 0.260 | 0.293 | 0.226 | 0.248 |
| Full | 0.168 | 0.223 | 0.223 | 0.237 | 0.165 | 0.201 | 0.212 | 0.222 |

Considering the generated systems, several observations can be made. Hardware devices communicating with software tasks tend to generate small subgraphs of tasks reliant on the raw data produced by these devices, since they encode some assumptions about the underlying hardware. These subgraphs appear in many solutions with identical structure, assigned to a single computer minimizing bandwidth consumption. In every mission requiring the target tracker, tasks identifying and localizing the target are assigned to the computer connected to the required sensor data, with the same structure of inputs and outputs between them. The task representing pose estimation with GPS and odometry data, however, would only sometimes share assignments with the GPS antenna interface. These tasks may encode more assumptions about the underlying hardware into software interfaces, resulting in more rigid structures. Tasks operating at higher levels show more flexibility in assignment. This behavior is driven by zero cost local routing in Equation 22 and lack of load balancing in Equation 21. This cost function generates lower costs for densely packed tasks, which coupled with the hardware assumptions present in tasks, generates these repeated subgraph structures.

In these experiments, three IMUs are included with different costs and connections, but lacking any differentiation in performance characteristics. As a result, one option offers the lowest cost, and is used in every system requiring inertial measurements. This introduces a small single board computer in order to support a physical transport type unsupported by every other device ($I^2C$). The combination of a small computer and a low cost IMU was introduced to model a subsystem composed of parts readily available from a hobbyist outlet. Generally, solutions treat the combination of IMU and single board computer identically to how the parts were initially introduced (i.e. as a single purpose inertial measurement module). In a few cases, other tasks get assigned to the single board computer, exploiting under-utilized resources. Two tasks commonly transitioned to this computer occur in Figure 7, in which the Kalman filter, which consumes IMU data directly, and the RGB camera interface (and matching camera). These cases demonstrate the power of system synthesis, since resources are utilized in a rigorously optimal fashion, as opposed to following local decisions.

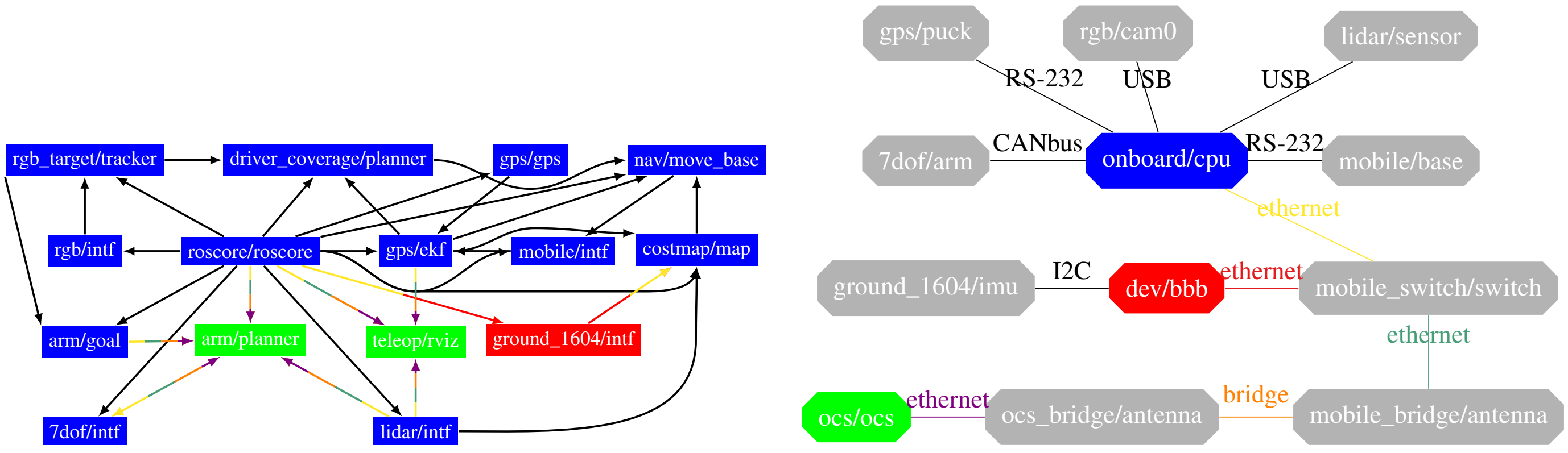

(a) Software graph

(b) Hardware graph

Fig. 7. Synthesized system for the complete configuration, when operating outside, during the day, with visible targets.

### *B. System Synthesis Scalability*

While the system synthesis problem is $\mathcal{NP}$-hard, the real world runtime for solving these problem informs the scale of problem which can be addressed. The complexity of the problem space admits many problem instances which do not reflect realistic design problems, such as a fully disconnected software graph or hardware components which can only produce a linear chain, which can result in unrealistic solution times. In order to provide empirical data for the problem scalability, we take one instance of the previously presented synthesis problem, and modify the number of components provided as input, and measure the time it takes to solve the resulting problem. Specifically, the complete configuration shown in Figure 7 serves as the base problem, with modules either added or removed to help introduce scaling. On the lower bound, the problem input is reduced to the minimal set of modules required to produce the solution, representing an under-constrained form of the underlying MRQARP problem. For scaling, differing numbers of additional components are included, with the average runtime over 5 trials reported in Table II.

The runtime numbers show that runtime increases largely derive from introducing additional hardware into the input. Doubling the number of tasks increases runtime, but by an order of magnitude less than doubling the devices. Similarly, doubling both shows runtime largely comparable to doubling the number of devices. Unsurprisingly, the examples including additional devices which introduce additional complexity to the underlying assignment or routing problems (i.e. computers which can execute tasks, or switches which can provide more potential routes) drive the increases in runtime.

### *C. Humanoid Disaster Response Robot Synthesis*

The development of robots to address novel real world scenarios pushes researchers to develop more complex systems. Competitions such as the DARPA Robotics Challenge serve to focus efforts on particular scenarios and encourage pushing the state of the art in fielded systems. These systems provide an ideal case study - reducing time spent diagnosing missing functionality or oversubscribed resources would lead to increased productivity and safety. Hard constraints imposed by competition rules and non-functional requirements derived from related efforts introduce additional need for expert knowledge to guide module capabilities. For instance, degraded communications between operator and robot constrains the hardware structure, as well as guiding software development of software to respect tighter bandwidth limits. Previous designs can provide insight into the new design space, as well as accumulated expert knowledge. The authors previously designed and deployed ESCHER in the DARPA Robotics Challenge, which provides a complex, well understood system which can be decomposed into components and reconstructed using system synthesis for comparison. As a custom humanoid, ESCHER's design included a wide variety of custom hardware and software developed and integrated into a single, functional whole. The hardware design included 33 degrees-of-freedom with both custom (e.g. linear series elastic actuators) and off-the-shelf (e.g. MultiSense sensor) devices. Furthermore, the software employs a mix of novel and open-source tasks covering three middlewares (ROS, LCM, and Bifrost), development shared between four teams [6], [37],

TABLE II
MEAN (N=5) SOLUTION TIMES (SECONDS) FOR DIFFERENT PROBLEM SCALES

| Devices | Tasks | Modules | Runtime(s) | Notes |
|---|---|---|---|---|
| 13 | 15 | 19 | 0.01172 | Minimal Input |
| 19 | 26 | 30 | 0.2465 | Original |
| 22 | 26 | 33 | 0.8026 | Original + 3 switches |
| 19 | 52 | 53 | 1.225 | Original + Duplicated Tasks |
| 25 | 26 | 36 | 2.784 | Original + 6 computers |
| 28 | 26 | 39 | 4.878 | Original + 3 switches, 6 computers |
| 38 | 26 | 36 | 12.93 | Original + Duplicated Devices |
| 38 | 52 | 60 | 13.7 | 2x Copy |
| 57 | 78 | 90 | 578.6 | 3x Copy |

[49] and totaling over 1.7 million lines of code. ESCHER's design was performed manually for the competition, providing a real world baseline for comparison.

Each design approach operates with differing levels of freedom in this comparison. Resource requirements, the organization of functionality into discrete devices or tasks, and the approach to meeting system requirements can evolve throughout the original development process, which cannot be captured with an *ex post facto* application of synthesis. Applying system synthesis on a completed design can provide insight into only the aspects of the design over which it operates - device and task interconnections, execution assignments, etc. Many of the decisions available during development cannot be considered within this framework in a single problem instance. This comparison focuses on providing some sense of how system synthesis compares to human effort in the later stages of integration and field experiments, in which both system synthesis and human experts perform equivalent tasks in design. In order to accomplish this, the final design of ESCHER is decomposed into a set of modules ($D = 18, P = 36, N = 12$) based on how functionality was organized in the implemented system, and used to generate and solve a full system synthesis problem instance. The components in these modules are modeled based on the known structure from the manual design, with resource usage derived from the operation of the robot. Note that system synthesis does not have access to the original structure, and must generate it *de novo*.

The automated solution (Figure 9) compares favorably with the manual result (Figure 8). The automated design eliminates two of the six computers, resulting in a more compact hardware design. The manual design included four onboard computers named sense, think, act, and dream. A fifth computer, known as archangel, off-board the robot but not suffering from degraded communications, managed the network connection. Manual task assignments mirrored the names: motion-related tasks to act, perception related tasks to sense, higher level cognitive functions to think, network management to archangel, and the remaining tasks to dream. In the synthesized system, these groups are approximately reproduced. The differences can be summarized as placing tasks closer to necessary inputs: tasks related to mid-level perception were assigned alongside the planning tasks which using their outputs, while network management tasks were moved alongside tasks generating data to transmit across the bridge. Localization and networking tasks handling control messages were assigned alongside connected motion tasks. It is interesting to note that while the software graph remains unchanged between two designs (intuitively, as the synthesis of a deconstructed design, the software structure requires the same flow of data), the hardware and assignment components change dramatically. This highlights one of the critical capabilities for integrating both structure and assignment into a single problem - rigidity in subsystems can appear to limit apparent design choices, when in fact flexibility extends beyond that subsystem. System Synthesis uses two fewer computers, rearranges tasks to use less bandwidth, producing an intuitively more efficient design with a mean solution time $0.410s$ ($n = 5$), well within design-time operation.

## V. Discussion & Conclusion

System synthesis captures several aspects of robot design, integrating them into a unified framework for ensuring a functional result. Introducing higher level concepts of functionality and context-awareness provide mechanisms for encoding expert knowledge about requirements at the systems engineering level, and restrictions on valid systems on the technical level, while still enforcing constraints on computing resources. These two features provide the mechanism for down-selecting core components to guide system construction. The other novel constraints introduced define a set of conditions necessary for components to operate, and that the resulting system provides a consistent solution. Integrating these constraints with the assignment and routing problem ensures that the fine grained details of constructing a viable system are handled as well. These features extend the underlying problem to provide a global solution from high level functional requirements to low level assignment concerns.

The case studies demonstrate the capabilities of this system in generating modern fielded robots. Demonstrated performance meets the design-time analysis role suggested in these experiments, automating a previously manual effort. Traditional approaches to designing computational systems for robots involves laborious trial-and-error efforts, or over-engineering the hardware aspect to ensure many viable solutions exist, without a rigorous framework for defining optimality. Automated system synthesis compares very favorably to these approaches, ensuring an optimal solution while providing a framework for quantifying several aspects of the design.

### A. Future Work

Posing system synthesis as an integer program results in an $\mathcal{NP}$-hard problem to find a globally optimal solution. Switching to approaches capable of quickly finding approximately optimal solutions raises the possibility of extending the range of problems which can solve in an online scenario. Extending this approach to address hardware fault recovery, dynamic operational contexts or mission requirements offers the ability to be more resilient in the face of failures. Dynamically altering hardware and software as requirements and context change can offer greater adaptability, allowing components developed to address specialized scenarios to be used when necessary. Generally, online execution of system synthesis can enable reasoning about interactions between a system's structure and the environment, potentially enabling a wide variety of new capabilities.

Another avenue of work introduces reasoning about the physical relationships between elements. Hardware connections currently serve to transfer data, though in reality can also transfer power, add mass, or enforce mechanical relationships. Generalizing hardware connections can allow reasoning about additional hardware constraints such as power and mass budgets. Consideration of mechanical relationships can improve expressiveness in synthesizing multi-robot systems, as well as enabling more expressive constraints on mechanical systems (e.g. ensuring an IMU is rigidly mounted, or selecting mobility elements based on environment).

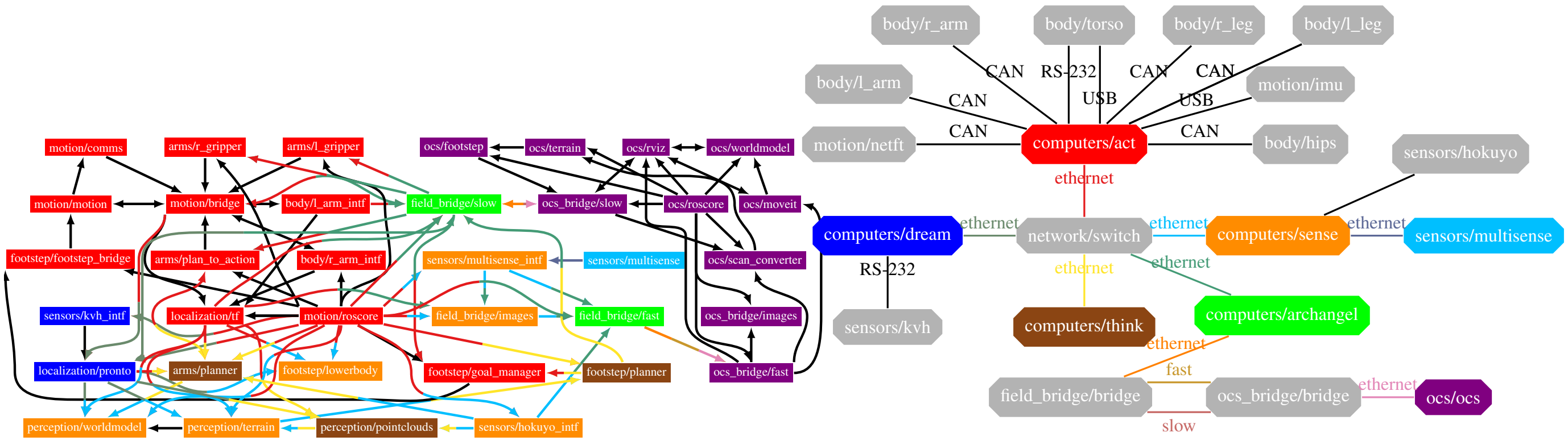

(a) Software Graph (b) Hardware Graph

Fig. 8. Manual ESCHER system design.

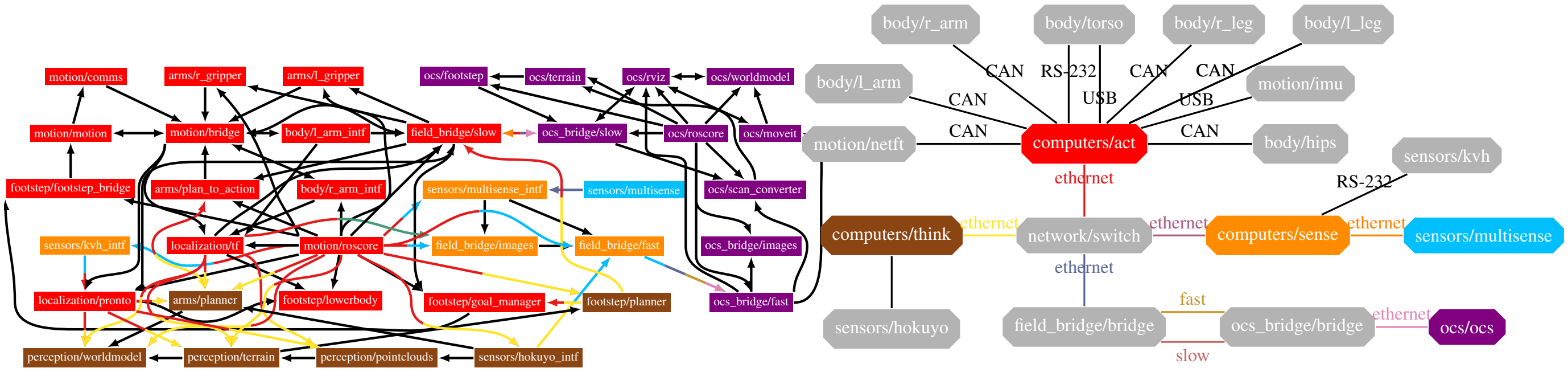

(a) Software Graph (b) Hardware Graph

Fig. 9. Synthesized ESCHER system design.